\documentclass{article}
\usepackage{ijcai24}

\usepackage{times}
\usepackage{soul}
\usepackage{utfsym}
\usepackage{url}
\usepackage{amsfonts}
\usepackage[hidelinks]{hyperref}
\usepackage[utf8]{inputenc}
\usepackage[small]{caption}
\usepackage{graphicx}
\usepackage{amsmath}
\usepackage{amsthm}

\usepackage{booktabs}
\usepackage{algorithm}
\usepackage{enumitem}
\usepackage{algorithmic}
\usepackage[switch]{lineno}
\usepackage{multirow}
\usepackage{multicol}
\usepackage{subcaption}

\title{Predictive Modeling with Temporal Graphical Representation on Electronic Health Records}

\author{
Jiayuan Chen
\and
Changchang Yin\and
Yuanlong Wang\And
Ping Zhang\thanks{The corresponding author. }\\
\affiliations
The Ohio State University\\
\emails
\{chen.12930, yin.731, wang.16050, zhang.10631\}@osu.edu
}

\begin{document}

\maketitle

\begin{abstract}
Deep learning-based predictive models, leveraging Electronic Health Records (EHR), are receiving increasing attention in healthcare. An effective representation of a patient's EHR should hierarchically encompass both the temporal relationships between historical visits and medical events, and the inherent structural information within these elements. Existing patient representation methods can be roughly categorized into sequential representation and graphical representation. The sequential representation methods focus only on the temporal relationships among longitudinal visits. On the other hand, the graphical representation approaches, while adept at extracting the graph-structured relationships between various medical events, fall short in effectively integrate temporal information. To capture both types of information, we model a patient's EHR as a novel temporal heterogeneous graph. This graph includes historical visits nodes and medical events nodes. It propagates structured information from medical event nodes to visit nodes and utilizes time-aware visit nodes to capture changes in the patient's health status. Furthermore, we introduce a novel temporal graph transformer (TRANS) that integrates temporal edge features, global positional encoding, and local structural encoding into heterogeneous graph convolution, capturing both temporal and structural information. We validate the effectiveness of TRANS through extensive experiments on three real-world datasets. The results show that our proposed approach achieves state-of-the-art performance.
\end{abstract}

\section{Introduction}


Electronic Health Records (EHR) play a crucial role in healthcare by enhancing the quality of care and improving the efficiency and accuracy of medical decisions. In recent years, with the advancement of artificial intelligence technologies, many deep learning models based on EHR data have been extensively explored and applied in predictive modeling, such as patient risk assessment \cite{10.1145/3219819.3219904,hitanet,graphcare,wang2024multimodal}, prediction of drug interactions \cite{leap,SafeDrug}, and support for clinical decision-making \cite{yin1,yang2023transformehr,chen2024spatial}.

Predictive modeling on EHR data is proposed to utilize clinical events from patients' historical visits, such as diagnoses, procedures and medications to forecast their future health status. These clinical events are typically represented as a set of medical codes. To enhance the quality and accuracy of medical predictive modeling, models need to learn three hierarchical representations in EHR: the \textbf{patient-level}, \textbf{visit-level}, and \textbf{(medical) code-level}. Specifically, the patient level representation should capture the temporal relationships between historical visits to build a comprehensive health profile of the patient. Then, at the visit level, the model analyzes the various medical events encountered in each visit to understand the specific circumstances of each visit. Finally, at the medical code level, the model needs to examine the characteristics of individual medical events and  their higher-order patterns across historical visits, as well as the correlations between different medical events.



\begin{figure}[t]
    \centering
    \includegraphics[width=0.48\textwidth]{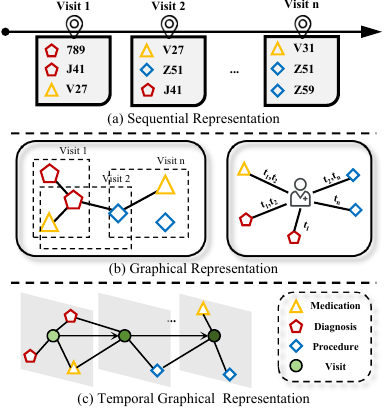}
    \caption{Different EHR representation methods. (a) An example of sequential representation. (b) Examples of graphical representation, which include graphs at the visit level and the patient level. (c) The temporal graphical representation proposed in this paper.}
    \label{fig:0}
\end{figure}

Patients' longitudinal EHR, as a series of successive visits, can be naturally conceptualized as a sequence, as illustrated in \autoref{fig:0}a. Current literature extensively employs sequence-based models, such as RNN-based \cite{RETAIN,stagenet} and Transformer-based \cite{hitanet,yang2023transformehr}, to track state changes across a patient's historical visits. 
These models excel in extracting temporal information at the visit level. However, they fall short in discerning structured relationships between medical events within visits due to weak temporal associations among these events. In contrast, graph-based models, another prominent category of EHR models, are adept at understanding relationships between medical codes in individual visits, especially regarding their structured connections. Some models \cite{GRAM,KAME,cgl,graphcare} enhance medical concept representations by integrating elements like medical ontology trees or subgraphs from medical knowledge graphs (KGs). As shown in Figure 1b, other approaches conceptualize EHR as assorted graphs,  applying graph-based models to extract structured information. For instance, \cite{GCT} has introduced a Graph Convolutional Transformer, aimed at deciphering the graphical structure of medical codes. While graph-based models are proficient in learning enhanced visit-level representations through structured features, they face challenges in integrating temporal data. DDHGNN \cite{DDHGNN} tries to remedy this by embedding dynamic temporal edges in heterogeneous patient graphs, achieving improved predictions but lacking an explicit focus on patient state changes over historical visits. This leads to a pivotal question: Is there a model capable of concurrently capturing all hierarchical latent relationships within EHRs, thereby achieving superior patient, visit, and medical code level representations

To address this challenge, in this paper, we model EHR as a novel temporal heterogeneous graph,  as depicted in the \autoref{fig:0}c. The visit-level and medical code-level representations are learned through dedicated medical event nodes and time-dependent visit nodes, respectively. The patient graph is designed to model intrinsic graphical
relationships via edges connecting medical event nodes to visit nodes. Additionally, it facilitates the propagation of temporal relationships reflective of changes in patient state through temporal edges interlinking these visits. Building upon the patient graph representation,  we introduce the Temporal gRAph traNSformer (\textbf{TRANS}), a novel model designed to deeply unravel these complex relationships. In particular, our temporal convolution merges temporal embedding with heterogeneous message passing. This fusion allows for the extraction of graph-structured features from medical events within individual visits, while also capturing the complex, higher-order temporal dependencies that span across historical visits. 
We enhance the learning of graphical features by integrating local structural encoding and global positional encoding into the medical event node representations, facilitated through meta-paths. Additionally, a patient explainer has been developed to ensure that our predictions are interpretable. We highlight the contributions of this work as follows:
\begin{itemize}
\item We propose a novel EHR representation method, transforming patient EHR into a heterogeneous graph with distinct sequential visit nodes and medical event nodes. This approach adeptly captures both structural relationships among medical events and temporal dependencies across historical visits.
\item A novel model TRANS with temporal graph convolution and the structural and positional encoding of heterogeneous graphs is proposed to extract the temporal graphical representation of a patient's graph, leading to interpretable predictions. 
\item  We conduct extensive experiments on three real-world datasets to demonstrate that both the constructed graph and the TRANS model outperform baseline methods, demonstrating the effectiveness of our approach to healthcare predictive modeling.  The implementation code can be found at Github\footnote{https://github.com/The-Real-JerryChen/TRANS}.
\end{itemize}

\section{Preliminary}
In this section, we provide some background knowledge about EHR data and then formulate the diagnosis prediction task. 
\subsection{Medical Codes} 
Each medical event can be represented as a set of unique medical codes, denoted as $ \mathcal{C}=\{c_1, c_2, \ldots, c_{|\mathcal{C}|}\}$, where $|\mathcal{C}|$ is the size of unique medical codes. Each code may represent a diagnosis, procedure, or medication. In this paper, we abstractly refer to these events as $e$.

\subsection{EHR Data} 
The EHR of each patient is represented as a time-sequenced array of medical events, encompassing diagnoses, procedures, and medications.  Specifically, the historical visit information of patient \(p\) can be represented as $\mathbf{X}^{p}=\left[\mathbf{x}_1, \mathbf{x}_2, \cdots, \mathbf{x}_T\right]$, where the $t$-th visit is denoted by a multi-hot vector $\mathbf{x}_t \in\{0,1\}^{|C|}$, $ t =1, \dots , T^p$ indicates the index of visit, and $T^p$ denotes the total visit times of patient $p$.
The $i$-th element of one visit vector is set to 1 if it contains the medical code $c_i$. To simplify, we explain the algorithms for one patient, omitting the superscript $p$ when clear.

\subsection{Sequential Diagnosis Prediction} 
The goal of predictive modeling is to predict the label at each time stamp or at the end of the visit sequence. In this study, we focus on the task of next-visit diagnosis prediction. It involves predicting  diagnosis codes $\mathcal{C}_{diag.}\in\{0,1\}^{|\mathcal{C}|}$ based on the medical events from historical visits as well as the medication codes and procedure codes for the current visit.

\section{Methodology}
\begin{figure}[t]
    \centering
    \includegraphics[width=0.48\textwidth]{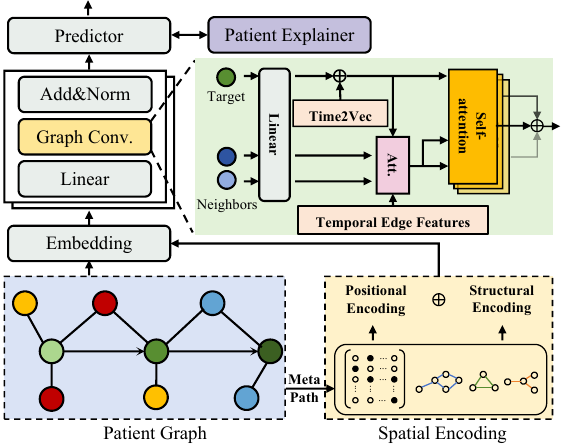}
    \caption{The overall framework of TRANS. 
The input EHR data is first constructed into a heterogeneous graph containing visit nodes and medical event nodes. Then the node features are mapped to an embedding space and combined with structural and sequential information. Subsequently, a temporal graph transformer is used to aggregate the node information. Finally, the features of the last visit node are input into a predictor.}

    \label{fig:enter-label}
\end{figure}

In this section, we first introduce the method for temporal graph construction, followed by details of the proposed model TRANS. \autoref{fig:enter-label} shows the overall architecture of TRANS, which is composed of three main modules: (1) heterogeneous temporal message passing;
(2) spatial encoder; and (3) patient graph explainer.
\subsection{Graph Construction}
To capture the structured information of potential disease development trends, treatment pathways, and other temporal graphical relationships within the patient EHR, we construct a heterogeneous patient graph $G=\left(\mathcal{N}, \mathcal{E}\right)$ for each patient, as illustrated in Figure 2,
 where $\mathcal{N}$ and $\mathcal{E}$ are sets of nodes and edges, respectively. The heterogeneous graph comprises four types of nodes: diagnoses, procedures, medications, and timestamped visit nodes: $\mathcal{N} = \mathcal{N}_e  \cup \mathcal{N}_v$. where $|\mathcal{N}_v| = T, |\mathcal{N}_e| = |\mathbf{e}|$. Here, $\mathbf{e}$ represents medical events that include diagnosis, medication, and procedure. The edge set contains two types of edges: (1)  undirected edges between medical codes and visits when a visit includes medical events $\mathcal{E}_{ev} = \{(u_1,u_2):\forall u_1\in \mathcal{N}_e, \forall u_2\in \mathcal{N}_v\}$; (2) directed edges between visits in chronological order $\mathcal{E}_{vv} = \{(u_t,u_{t+1}\}):\forall u\in \mathcal{N}_v, 0\leq t  < T\}$. 

To the best of our knowledge, our graph construction differs from existing methods \cite{GCT,DDHGNN,graphcare} of EHR representation learning. We explicitly represent each visit with time-dependent visit node. This approach decouples the representations of medical codes and the historical visits, allowing the embedding of clinical event nodes to focus on the characteristics of the codes themselves, while using the embedding of visit nodes to capture changes in the patient's state. Additionally, we incorporate temporal embedding on the edges between medical codes and visits, as well as between successive visits with different time indexes.

\subsection{Temporal Heterogeneous Message Passing}
To effectively gather temporal information and advanced connections between medical events across historical visits, we propose a tailored temporal-aware message-passing mechanism for heterogeneous graphs. This system not only propagates medical code information into patient visit representations but also subtly connects various medical events through the visit nodes. We represent the features of medical code node and visit node outputted by the $(l)$-th layer of the network as $\mathbf{h}^{(l)}_{e}\in \mathbb{R}^d$
and $\mathbf{h}^{(l)}_{v}\in \mathbb{R}^d$, respectively. They also serve as the input for the subsequent $(l+1)$-th layer. We denote linear mappings as $\text{L}(\cdot)$, with each instance representing a distinct mapping.

\subsubsection{Temporal Embedding}
The temporal embedding in TRANS consists of two parts: the time encoding of the visit nodes; and the temporal encoding of edge features. We employ Time2Vec \cite{time2vec} to encode the absolute times of the visits, normalized for each patient, into vectors. This time-aware vector is then concatenated with the initial features of the visit node $\mathbf{h}^{(0)}_{v}$, thereby incorporating information about the time intervals between different visits.

Time2Vec is suitable for handling real-world timestamps, such as specific visit times, while for the indexed time of edges between medical events and visits, we use functional time encoding. Inspired by \cite{iclr2020}, we use $\mathcal{T}(t)$  to encode visit index $t$ into a continuous vector, which is then regulated by a linear layer to obtain the time factor $\alpha_t$.
\begin{equation*} 
    \mathcal{T}(t) =\sqrt{\frac{1}{d}}\left[\cos \left(\omega_1 t\right), \sin \left(\omega_1 t\right), \ldots, \cos \left(\omega_d t\right), \sin \left(\omega_d t\right)\right],
\end{equation*}
where $\omega_1,\dots,\omega_d$ are learnable parameters. And $\alpha_t =\text{L}\left(\mathcal{T}\left(t\right) \right).$ We separate the temporal vector from node features, treating it as an edge type, to maintain focus on the nodes' semantic information while infusing temporal data during message propagation into visit node features. Details can be found in the appendix.

\subsubsection{Heterogeneous Message Passing}
The entire message-passing process can be formalized into two stages: aggregation and updating. The first step is to aggregate the information of neighboring nodes. Specifically, we use an attention mechanism to weigh and integrate the features of neighboring nodes and concatenate the output from multi-head attention to obtain the final message. Taking as an example the process of propagating the features of neighbor nodes $e_t \in N(v_t)$ to the visit node $v_t$, the i-th head attention is as follows:
\begin{align}
&\mathbf{q}_v = \text{L}\left(\mathbf{h}^{(l)}_{v}\right), \quad \mathbf{k}_{e_t} = \text{L}\left(\mathbf{h}^{(l)}_{e_t}\right) \notag\\
&\text{Att}^{i}(e_t, v_t)  =   \frac{\alpha_t \cdot\left(\mathbf{q}_v \mathbf{W}^{\mu(e)} \mathbf{k}_e^\top\right)}{\sqrt{d}},
\end{align}
where $e_t$ represents the medical event node such as diagnosis, medication, and procedure contained in $t$-th visit as well as previous visit node $v_{t-1}$. The target node for message propagation is $t$-th visit node $v_t$.
We use weight matrix $\mathbf{W}^{\mu(e)}\in \mathbb{R}^{d\times d}$ to encode different types of medical events, where $\mu(e)$ represents the type of the node $e_t$, such as diagnosis, procedure, or medication.
After obtaining the attention scores for different neighboring nodes, we combine them with the mapped neighbor features to complete the entire aggregation process.  We formalize it as follows:
\begin{equation}
   \widetilde{\mathbf{h}}^{i}_v = \underset{\forall e_t \in N(v_t)}{\text{Softmax}}\left(\text{Att}^{i}(e_t, v_t)\right) \cdot \text{L}(\textbf{h}^{(l)}_{e_t}),
\end{equation}
where $N(v_t)$ denotes neighbors of $v_t$, and $ \widetilde{\mathbf{h}}^{i}_v$ represents the message passed to the visit node  by the $i$-th head attention. Following the acquisition of aggregated information, the next stage is to combine this aggregated information with the target node's ego information to update the representation of the target node, as follows:
\begin{align}
&\textbf{h}^{(l+1),i}_{v} = \gamma\cdot\text {L }\left(\sigma\left(\widetilde{\mathbf{h}}^{i}_v\right)\right)+(1-\gamma)\cdot\textbf{h}^{(l)}_{v}, \notag\\
&\textbf{h}^{(l+1)}_{v} =\underset{i\in[1,h]}{\|} \textbf{h}^{(l+1),i}_{v}
\end{align}
where $\underset{i\in[1,h]}{\|}$ denotes concatenating the outputs of multiple heads, $\gamma$ is the coefficient for the skip connection and $\sigma$ represents activation function. Similar to the network structure in \cite{gcnii} , it can be either a learnable parameter or a hyper-parameter. Leveraging heterogeneous message passing, we integrate the features of medical events in the visit, along with the graphical structure information and historical visit information, into the current visit node, thereby updating the patient's representation.

\subsection{Spatial Encoding}
 We integrate global positional encoding and local structural encoding into the initial feature of medical event nodes by concatenating them. This approach effectively captures structured features from patients' historical visits, which is crucial for illuminating key patterns like diagnosis sequences and treatment regimes, essential for accurately modeling disease stages. Such an integration enables us to discern intricate relationships in medical event co-occurrences and disease trajectories, thereby enhancing our predictive accuracy for a patient's future health status.

Local structural encoding $\mathcal{S}(e)$ is designed to capture local structural information in the graph based on random walks. Positional encoding $\mathcal{P}(e)$ adds a global positional encoding based on the Laplacian eigenvectors of the patient graph. However, in heterogeneous graphs, unlike homogeneous graphs \cite{GPS}, direct computation of certain properties is not straightforward, owing to the complexity introduced by diverse types of nodes and edges. To navigate this complexity, we employ medical event meta-paths, offering a structured and meaningful method to traverse the heterogeneous graph. For instance, in constructing meta-paths for medication nodes, visit nodes are used as intermediaries to connect multi-hop neighbors, forming a sequence represented as:
\begin{equation*}
(N_m, N_v)-(N_v, N_d)-(N_d, N_v)-(N_v, N_m),
\end{equation*}
where $N_m$, $N_v$, and $N_d$ represent medication, visit, and diagnosis nodes, respectively, and $-$ denotes the pseudo edge. This meta-path construction results in an adjacency matrix $\Tilde{\mathbf{A}}\in \mathbb{R}^{|N_m| \times |N_m|}$, where $|N_m|$ is the number of unique medication nodes within graph. 

Once the meta-paths are established, we can compute structural embedding based on the corresponding adjacency matrix $\Tilde{\mathbf{A}}$. And for $\mathcal{P}(e)$, we construct a Laplacian matrix and compute its top $k$ eigenvectors to represent the spatial information of nodes within the graph. $\mathcal{P}(e)\in \mathbb{R}^{k}$ for node $e$ is expressed:
\begin{equation}
\mathcal{P}(e) = [\mathbf{v}^{(1)}_e, \mathbf{v}^{(2)}_e,\dots, \mathbf{v}^{(i)}_e, \dots,\mathbf{v}^{(k)}_e],
\end{equation}
where each $\mathbf{v}^{(i)}_e$ is the element corresponding to node $e$ in the $i$-th eigenvector. For structural encoding $\mathcal{S}(e)\in \mathbb{R}^{k}$, we calculate $k$-step random walk matrices $\Tilde{\mathbf{W}}^{k}$ based on the adjacency matrix $\Tilde{\mathbf{A}}$, and extract the cumulative probabilities of node $e$ connecting to itself at each step to form $\mathcal{S}(e)$ as :
\begin{equation}
\mathcal{S}(e) = [ \mathbf{d}^{(1)}_e, \mathbf{d}^{(2)}_e, \dots,  \mathbf{d}^{(i)}_e, \dots, \mathbf{d}^{(k)}_e ],
\end{equation}
where $\mathbf{d}^{(i)}_e$ represents the diagonal element of node  $e$ after the $i$-th step of random walk. This approach not only enhances the representational ability of the graph but also provides a richer context for the deep analysis of complex interactions between different types of medical events.

\subsection{Patient Explainer}
Based on the patient graph, we are able to predict the future development of a patient's status using their historical diagnosis and medical information. However, the decision-making process of GNNs also requires intuitive and interpretable analysis to increase the credibility of decisions and to assist doctors in better understanding them.  
Therefore, we have incorporated a heterogeneous graph explainer module into our model. By masking certain medical event nodes, the explainer aims to identify a significant subgraph $G_s$ crucial for the model's predictive results. Utilizing mutual information (MI) as the criterion, this process can be expressed as follows:
\begin{equation}
\max _{G_s} \text{ MI}\left(\hat{Y}, G_s\right)
\end{equation}
where $\hat{Y}$ is the prediction of the TRANS model. The obtained subgraph contains the most influential medical events in the EHR for predicting the patient's status. We will provide a detailed introduction to this module in the appendix

\section{Experiments}

\subsection{Experimental Setting}
\subsubsection{Datasets}
We conduct experiments on the MIMIC-III \cite{mimic3}, MIMIC-IV \cite{mimic4}, and MarketScan Commercial Claims and Encounters (CCAE) \cite{ccae} datasets. MIMIC-III and MIMIC-IV are two publicly available medical databases released on PhysioNet. Following \cite{KAME}, we filtered the original datasets to include patients who made at least two visits. The CCAE\footnote{Available at https://www.merative.com/real-world-evidence.} dataset is a healthcare database, which contains patients with longer historical visit information.  The goal of the diagnosis prediction task is to predict the diagnostic labels for the next visit. Similar to \cite{KAME}, we chose to predict the category of the diagnosis, which is the CCS (Clinical Classifications Software) code corresponding to the ICD code. We use PyHealth \cite{pyhealth2023yang} for parsing and preprocessing the dataset. Table \ref{tab:my_label} presents some statistics of the processed datasets. Details on \textbf{dataset processing} can be found in the Appendix.


\begin{table}[t]
    \centering
    \resizebox{1.0\linewidth}{!}{
    \begin{tabular}{l|ccc}
    \toprule
      Dataset   & MIMIC-III &MIMIC-IV &  CCAE \\
         \midrule
     \# of patients & 5,449 & 14,155 & 565,897\\
\# of visits  & 14,141 & 42,053 & 5,355,668 \\
Avg. \# of visits per patient & 2.60 & 2.97 & 9.46\\
Max \# of visits per patient & 39 &70& 257\\
\midrule
\# of unique diagnoses & 3,874 & 11,225 & 8,751\\
\# of unique procedures &1,412 & 8,352 & 11,358\\
\# of unique medicines&193 &196 & 200\\
\midrule
\# of CCS codes & 264 & 275& 280\\
         \bottomrule
    \end{tabular}
    }
    \caption{ Statistics of processed datasets. }
    \label{tab:my_label}
\end{table}

\subsubsection{Baselines}
We evaluate our model against the following baselines:
\begin{itemize}[leftmargin=0pt]
\renewcommand{\labelitemi}{}
\item \textbf{Sequence-based Models}. We use four sequence models based on different backbones. The Transformer \cite{Attention} consists of encoders stacked based on the self-attention mechanism. RETAIN \cite{RETAIN} is an RNN-based model that integrates a Reverse Time Attention mechanism. StageNet \cite{stagenet} is enabled by a stage-aware LSTM module and a stage-adaptive convolutional module. HiTANet \cite{hitanet} proposes a hierarchical time-aware Transformer model.
\item \textbf{Graph-based Models}. We compared our model with three graph-based models: KAME \cite{KAME}, GCT \cite{GCT}, and DDHGNN \cite{DDHGNN}. KAME incorporates medical ontology knowledge of ICD codes into the sequence model. GCT utilizes graph convolutional networks to learn the graphical structure in EHR. We extend it by concatenating a transformer, making it suitable for sequential input tasks. DDHGNN constructs EHR as heterogeneous graphs of patients and employs dynamic graph convolution to learn patient's features. Additionally, the results of some models \cite{hgt,anonymous2023graph} can also be regarded as ablation studies for our proposed model.
\end{itemize}
The \textbf{experimental setup} and \textbf{implementation details} can be found in the Appendix. %

\subsubsection{Evaluation Metrics}
We use two evaluation metrics \cite{KAME} for the diagnosis prediction task: visit-level precision@k and code-level accuracy@k. Visit-level precision@k is defined as the number of correct medical codes within the top k predictions, normalized by the minimum of k and the actual number of categories present in a visit. Code-level accuracy@k is the ratio of correct predictions to the total number of predictions made for individual codes. The value of k varies from 10 to 30, with higher values indicating better model performance, where visit-level precision assesses broader performance and code-level accuracy reflects more detailed accuracy.

\begin{table*}
\resizebox{1.0\textwidth}{!}{
  \begin{tabular}{cc|ccc|ccc}
  \toprule
\multirow{2}{*}{Dataset} & \multirow{2}{*}{Model} & \multicolumn{3}{c}{Visit-Level Precision@$k$} & \multicolumn{3}{c}{Code-Level Accuracy@$k$} \\

& & 10 & 20 & 30 & 10 & 20 & 30\\
\midrule
\multirow{9}{*}{MIMIC-III}&  Transformer &   $55.01\pm0.47$ & $62.51\pm0.37$ & $72.30\pm0.37$ & $39.99\pm0.37 $ & $58.40\pm0.44$ & $69.73\pm0.39$  \\
 & RETAIN &  $55.92\pm0.31$ & $63.51\pm0.29$ & $ 73.15\pm0.25$& $40.56\pm0.37$ & $59.27\pm0.35$ & $70.55\pm0.29$        \\
 & KAME & $52.94\pm0.41$ & $60.76\pm0.39$  & $70.74\pm0.31$ & $39.56\pm0.42$ & $57.48\pm0.28$ & $68.66\pm0.20$  \\
 & StageNet &  $51.49\pm0.43$ & $59.57\pm0.40$ & $69.33\pm0.30$ & $38.08\pm0.33$ & $55.72\pm0.32$ & $66.61\pm0.25$ \\
 & HiTANet & $56.33\pm0.38$ & $63.66\pm0.35$&$\underline{73.38\pm0.29}$ & $40.95\pm0.30$ & $59.87\pm0.27$ & $\underline{70.68\pm0.21}$\\
 & GCT & $\underline{56.95\pm0.28}$& $\underline{63.78\pm0.26}$& $73.16\pm0.20$ & $\underline{41.01\pm0.29}$ & $\underline{59.94\pm0.25}$ & $70.66\pm0.17$\\
 & DDHGNN & $56.80\pm0.35$& $63.73\pm0.23$ & $72.99\pm0.19$ & $40.59\pm0.36$ & $59.90\pm0.27$ & $70.59\pm0.20$  \\
& TRANS (Ours)& $\mathbf{57.65\pm0.25}$ & $\mathbf{64.18\pm0.17}$ & $\mathbf{73.69\pm0.16}$& $\mathbf{41.30\pm0.28}$& $\mathbf{60.19\pm0.19}$ & $\mathbf{71.17\pm0.12}$\\
\midrule
 \multirow{9}{*}{MIMIC-IV}&  Transformer & $62.27\pm0.39$ & $65.98\pm0.25$ & $74.66\pm0.15$ & $40.32\pm0.33$ & $60.07\pm0.26$ & $71.62\pm0.17$ \\
 & RETAIN &  $60.79\pm0.33$& $65.20\pm0.20$ & $74.10\pm0.16$ & $39.18\pm0.32$ & $59.20\pm0.26$ & $71.04\pm0.20$  \\
 & KAME &  $61.16\pm0.40$ &  $64.55\pm0.33$  & $73.03\pm0.20$ & $39.42\pm0.41$ & $58.72\pm0.29$ & $70.01\pm0.19$      \\
 & StageNet & $57.89\pm0.44$ & $61.97\pm0.31$ & $71.21\pm0.22$ &$37.15\pm0.46$ & $55.87\pm0.30$ & $67.80\pm0.21$   \\
 & HiTANet & $63.01\pm0.30$ & $66.24\pm0.24$  &$75.01\pm0.19$ & $40.51\pm0.34$ & $60.04\pm0.25$ & $72.19\pm0.15$\\
 & GCT & $\underline{63.15\pm0.27}$ & $\underline{66.39\pm0.21}$ & $\underline{75.05\pm0.20}$& $\underline{41.00\pm0.30}$ & $\underline{60.30\pm0.24}$ & $\underline{72.23\pm0.21}$\\
 & DDHGNN &   $59.67\pm0.33$ & $64.61\pm0.26$  & $73.32\pm0.18$  & $39.11\pm0.19$ & $59.54\pm0.15$ & $70.16\pm0.13$   \\
& TRANS (Ours)& $\mathbf{65.68\pm0.22}$ & $\mathbf{68.99\pm0.20}$   & $\mathbf{77.12\pm0.15}$ & $\mathbf{42.83\pm0.20}$ & $\mathbf{63.04\pm0.18}$  & $\mathbf{74.05\pm0.13}$\\
\midrule
 \multirow{9}{*}{CCAE}&  Transformer &   $68.97\pm0.09$ & $81.47\pm0.08$ & $87.56\pm0.08$ & $65.86\pm0.10$ & $78.68\pm0.08$ & $86.55\pm0.08$  \\
 & RETAIN & $68.14\pm0.10$ &   $77.12\pm0.09$ & $87.14\pm 0.10$ & $65.60\pm0.08$ & $76.98\pm0.07$ & $85.92\pm 0.07$  \\
     
 & KAME & $71.69\pm0.12$ & $81.72\pm0.10$ & $86.87\pm0.08$ & $66.59\pm0.11$ & $78.50\pm0.12$ & $84.63\pm0.10$\\
 & StageNet &  $74.72\pm0.12$ &   $84.44\pm0.12$ & $88.86\pm0.08$ & $69.65\pm0.11$ & $81.22\pm0.10$ & $86.61\pm0.09$      \\
 & HiTANet &   $\underline{75.68\pm0.11}$  & $\underline{84.63\pm0.13}$ & $\underline{89.29\pm0.10}$ & $\underline{70.35\pm0.15}$ & ${81.21\pm0.11}$ & $\underline{86.99\pm0.11}$ \\  
 & GCT & $75.64\pm0.14$  & $84.57\pm0.12$ & $89.09\pm0.09$ & $70.32\pm0.14$ & $\underline{81.22\pm0.13}$ &  $86.75\pm0.09$\\
 & DDHGNN &   $74.49\pm0.15$ & $84.13\pm0.13$ & $88.39\pm0.13$ & $69.20\pm0.16$ & $81.07\pm0.15$ & $85.89\pm0.15$\\
& TRANS (Ours) & $\mathbf{81.81\pm0.07}$  & $\mathbf{89.29\pm0.05}$ & $\mathbf{92.82\pm0.06}$ & $\mathbf{76.47\pm0.07}$& $\mathbf{86.15\pm0.06}$ & $\mathbf{90.70\pm0.04}$\\
\bottomrule
  \end{tabular}

}

\caption{Results for the diagnosis prediction task on MIMIC-III, MIMIC-IV and CCAE dataset, where the best results are highlighted in \textbf{bold} and the second best scores are \underline{underlined}.}
  \label{tab:res}
\end{table*}
\subsection{Experimental Results}
\label{exps}
Table \ref{tab:res} demonstrates the performance of the proposed TRANS model and all the other baselines on three datasets, including MIMIC-III, MIMIC-IV, and CCAE. We ran 10 times for each model under random dataset splits and reported the mean and standard deviation of the results, where the best results are highlighted in \textbf{bold} and the second best scores are \underline{underlined}.

\noindent \textbf{Overall results.} Compared to baseline models, our proposed model TRANS demonstrated SOTA performance across all three datasets. Notably, the improvements on the MIMIC-IV and CCAE datasets were more pronounced than on MIMIC-III. On the other hand, compared to MIMIC-III, almost all models performed better on the MIMIC-IV and CCAE datasets. This outcome suggests that more comprehensive datasets can enhance model training outcomes.

\noindent \textbf{Sequence-based models.} As mentioned before, sequence models treat patients' historical visits as text, where visits and medical events correspond to sentences and words, respectively.  Table \ref{tab:res} shows that for the MIMIC-III and MIMIC-IV datasets, it's evident that the overall performance of sequence models is not as good as that of graph-based models. This may be attributed to the fact that in contrast to graph-based models like GCT and DDHGNN, sequence models are less effective at extracting structured information within a visit. Our proposed TRANS, which constructs a patient graph with temporally connected visit nodes, therefore achieves significant improvements. For the CCAE dataset, our model also outperforms those sequence model-based models, which further highlights the critical role of the graph module. 

\noindent \textbf{Graph-based models.} 
In baseline models like KAME, tree-structured information of ICD codes is incorporated into the sequence structure. However, this does not deeply relate to the patient's state, resulting in KAME's suboptimal performance on several datasets. On the other hand, GCT, DDHGNN, and our proposed TRANS model integrate a graph feature extraction module for visit data into the sequential network, leading to notable improvements on datasets with shorter historical sequences like MIMIC-III and MIMIC-IV. Yet, as Table \ref{tab:res} indicates, these graph-based models are outperformed by ours on the CCAE dataset. This disparity may stem from their limited focus on visit characteristics, essential for capturing state changes in longer sequences for precise diagnosis. HiTANe's good performance on longer sequence datasets also reflects its hierarchical approach to analyzing a patient's historical visit stages.

\subsection{Model Analyses}
\subsubsection{Ablation Study}
In this section, we conduct ablation studies to investigate the contributions of different components. Specifically, we removed three key parts of TRANS, namely Spatial Encoding (SE), Temporal Embedding (TE), and Sequence information in node feature initialization (Seq). Spatial Encoding includes both structural encoding and positional encoding. Temporal Embedding comprises the Time2Vec features of visit nodes and the temporal embedding of edge features in the graph. As shown in Table \ref{tab:ab}, the experimental results first reveal that each of these components contributes to improving the model's performance to some extent. Second, we observe that removing TE causes more performance degradation than removing SE, especially in the CCAE dataset where both Seq and TE play a more significant role. This suggests that graph convolution layers in TRANS are capable of capturing structured information to a certain degree, which overlaps with some features in SE. However, the inclusion of SE also results in better performance, indicating its important role.

\begin{table}[ht]
    \centering
    \resizebox{0.48\textwidth}{!}{
    \begin{tabular}{ccc|ccc}
    \toprule
       SE & TE & Seq & MIMIC-III & MIMIC-IV & CCAE \\
         \midrule
        \usym{2717}    & \usym{2717}&\usym{2717} & $55.03\pm0.35$ &$62.23\pm0.33$ & $68.80\pm0.14$\\
        \usym{2717}    & \usym{2717}& \usym{2713} & $55.92\pm0.36$ & $62.74\pm0.27$ & $74.14\pm0.12$ \\
        \usym{2713}    & \usym{2717}& \usym{2713} & $56.73\pm0.32$ & $63.97\pm0.27$ & $76.75\pm0.10$\\
        \usym{2717}    & \usym{2713}& \usym{2713} & $57.16\pm0.28$ & $65.48\pm0.24$ & $81.63\pm0.08$\\
        \usym{2713}    & \usym{2713}& \usym{2713} & $57.65\pm0.25 $& $65.68\pm0.22$& $81.81\pm0.07$ \\
         \bottomrule

    \end{tabular}}
             \caption{Variants of TRANS. We only report average precision of the visit-level Top@10 due to the space limitations.}
    \label{tab:ab}
\end{table}

\subsubsection{Graph Construction}
In this section, we delve into the distinctions among various graph construction methods. There are three notable approaches: (1) The first method, exemplified by GCT \cite{GCT}, prioritizes the structural relationships between medical events within individual visits, omitting temporal data. This method aggregates medical event nodes to create a patient representation. (2) The second approach, as introduced by DDHGNN \cite{DDHGNN}, conceptualizes EHR as dynamic graphs with patient nodes at their core. It infuses temporal information into the dynamic edges linked to medical events. (3) Our proposed method, the third approach, models EHRs utilizing both time-dependent visit nodes and medical event nodes. The comparative performance of these methods is showcased in \autoref{tab:res}, where our method demonstrates superior efficacy. 

\begin{figure}[t]
    \centering
    \begin{subfigure}[b]{0.18\textwidth}
        \centering
\includegraphics[width=0.9\textwidth,height=2.1cm]{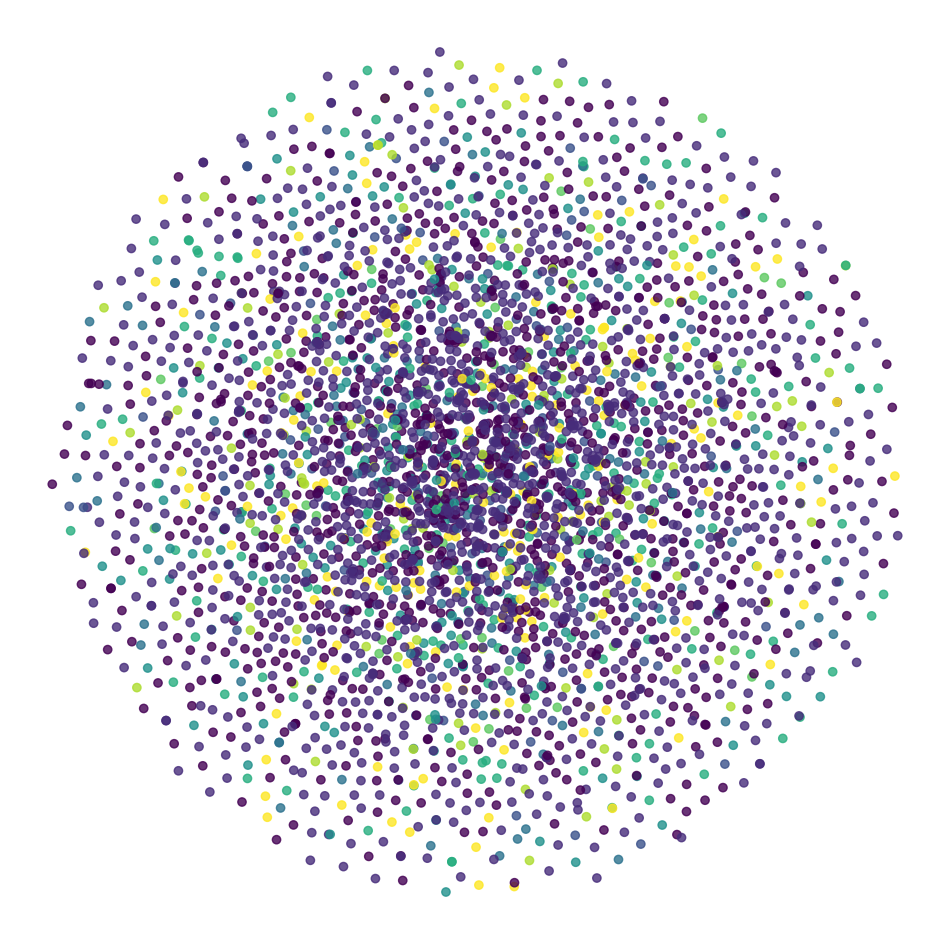}
        \caption{DDHGNN}
        \label{fig:ddhgnn}
    \end{subfigure}
    \begin{subfigure}[b]{0.18\textwidth}
        \centering
        \includegraphics[width=0.9\textwidth,height=2.1cm]{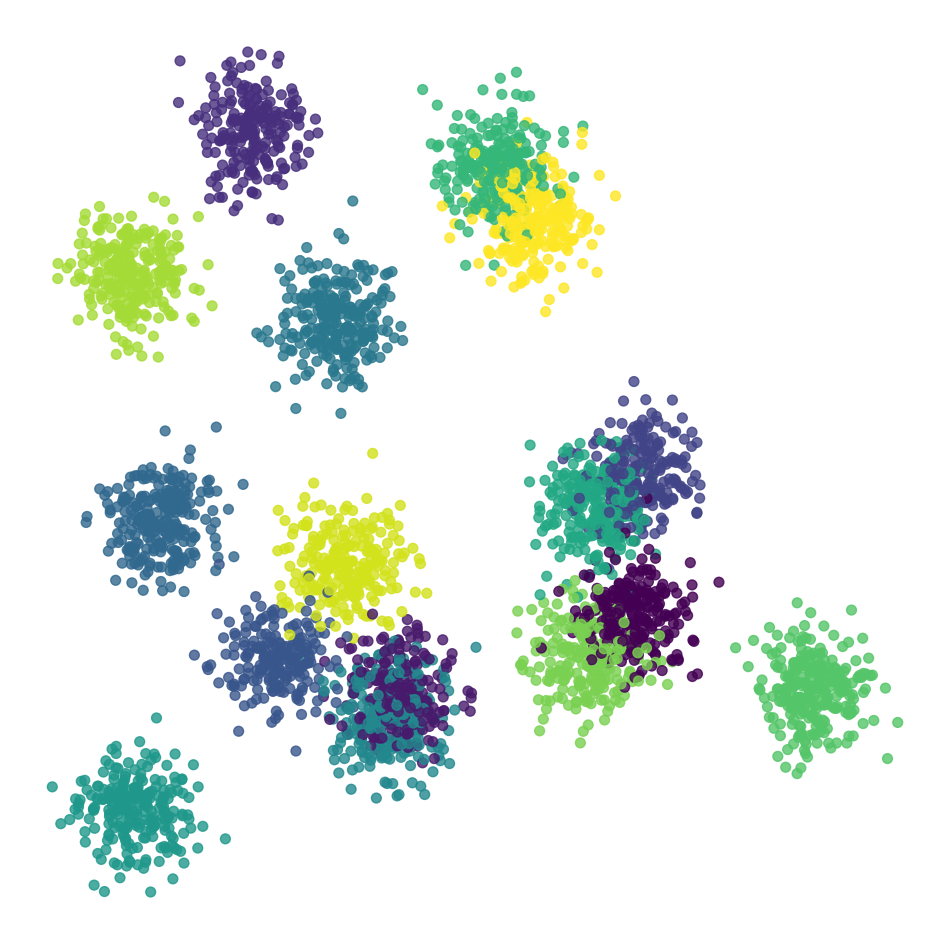}
        \caption{TRANS}
        \label{fig:ours}
    \end{subfigure}
    \caption{t-SNE Scatterplots of medical codes Learned by DDHGNN and TRANS on the CCAE dataset.}
    \label{tsne}
\end{figure}

DDHGNN's graph construction, while acknowledging historical visit temporal information, incorporates it solely into dynamic edge features. This leads to a conflation of medical event node representations with patient state data, diluting the focus on specific ICD code attributes. In our analysis of the CCAE dataset, we selected 4,000 codes that reflect a balanced distribution across disease categories. We visualized these codes in a 2D space using t-SNE (\autoref{tsne}), where different colors represent different categories. The clustering patterns of ICD code representations in our model (\autoref{fig:ours}) are notably distinct and category-specific, unlike the more chaotic representations observed in DDHGNN (\autoref{fig:ddhgnn}). This observation underscores the efficacy and superiority of our approach in graph construction.

\subsection{Graph Explainer}
We employ a patient graph explainer to enhance the interpretability of our model's predictions, which is crucial for dissecting the predictive role of each node, including medical event nodes like medications and diagnoses, and visit nodes. Leveraging our capability to discern the significance of various nodes, we have performed a detailed statistical analysis of key historical diagnoses within the CCAE dataset, which specifically pertains to patients with Chronic Obstructive Pulmonary Disease (COPD). For each COPD patient (labeled with CCS code `127'), we calculate the importance of each diagnosis occurrence and then average these values across the entire dataset. Our analysis of the diagnostic nodes related to COPD allows us to pinpoint several potential comorbidities. As depicted in \autoref{tab:c}, the explainer has identified a range of likely comorbidities, systematically categorized by disease type. Notably, these findings, encompassing prevalent complications such as Asthma, Bronchitis, and Hypertension, resonate strongly with established clinical knowledge\footnote{https://www.mayoclinic.org/diseases-conditions/copd/symptoms-causes/syc-20353679}, demonstrating the efficacy of the TRANS on predictive modeling.

\begin{table}[h]
    \centering
    \begin{tabular}{l}
        \toprule
        
         \textbf{Respiratory System Diseases}  \\
         \midrule
        J06 - Acute upper respiratory infections \\
        J20 - Acute bronchitis\\
        J45 - Asthma\\
        \toprule
        \textbf{Mental and behavioral disorders}\\
        \midrule
        F32 - Major depressive disorder\\
        F17 - Nicotine dependence\\
        \toprule
        \textbf{Circulatory System Diseases}\\
                 \midrule
        I10 - Essential hypertension\\
        E78 - Disorders of lipoprotein metabolism\\
             \bottomrule
    \end{tabular}
    \caption{Potential COPD comorbidities grouped by disease categories, as identified by the Patient Explainer.}
    \label{tab:c}
\end{table}


\section{Related Work} 
In this section, We review two main EHR representation methods: sequence-based and graph-based, along with their predictive modeling algorithms. 

\noindent
\textbf{Sequence-based models.} 
Sequence-based models utilize sequential models to capture temporal information in EHR. 
In sequence-based studies, patients' EHRs are represented a sequence of visit. Each visit contains a list of medical events (e.g., diagnosis codes and medications). These studies focus on extracting temporal or contextual features between visits.
For example, the classic model RETAIN \cite{RETAIN} introduces a attention-based combination of forward and backward visit-level representations to obtain more accurate predictions. Dipole \cite{Dipole} employs bidirectional recurrent neural networks combined with three distinct attention mechanisms for patient visit information prediction.  More recently, many works \cite{stagenet,hitanet,10020513,yang2023transformehr} have applied the powerful Transformer-based model adapted from natural language processing like BERT \cite{bert} to EHR, treating each visit or medical code as a token. 

\noindent
\textbf{Graph-based models.} Different from sequence-based models, graph-based models represent EHR in the form of graphs, which naturally model the inherent structural information among medical events.
For instance, many models leverage external medical KGs to improve medical code representation learning. To name a few, GRAM \cite{GRAM} utilizes medical ontologies and graph-based attention to develop strong medical code representations, while CGL \cite{cgl} utilizes disease domain knowledge and unstructured text. KerPrint \cite{KerPrint} proposed a time-aware KG attention method to understand knowledge decay over time. There is also considerable research focusing on the extraction of graph structural features within EHR. \cite{mime,GCT} explored the prior and hidden structured relationships between medical events. \cite{DDHGNN} expanded the graph from visit-level to patient-level, constructing a dynamic graph to represent EHR. However, existing graph-based models fail to simultaneously extract the temporal and structured relationships between visits.

\section{Conclusion}
In this paper, we investigate the three levels of relationships with EHR and the limitations of existing patient representation methods for predictive modeling. To capture both temporal and structural information about EHR, we propose a novel temporal heterogeneous graph to represent each patient's EHR. Moreover, we propose a TRANS model to learn the three-level relationships, which use temporal graph transformer to extract temporal and structured information from the graph. We validated our methods and the TRANS model through experiments on three EHR datasets, demonstrating superior performance in diagnosis prediction. Utilizing KGs in healthcare AI is also a crucial research area, particularly for improving the effectiveness of predictive models in the field. We will explore the expansion of our models through the incorporation of KGs in the future work.

\section*{Ethical Statement}
The datasets used in the paper are from PhysioNet (MIMIC-III and MIMIC-IV) and Merative (CCAE), which are fully HIPAA-compliant de-identified, and have very minimal risk of the potential for loss of privacy. Moreover, Per the DUAs, all users to access the data will need to take full research, ethics, and compliance training courses. Thus, potential privacy and security risks would be eliminated and/or mitigated.

\section*{Acknowledgments}
This work was funded in part by the National Science Foundation under award number IIS-2145625, by the National Institutes of Health under award number R01GM141279, and by The Ohio State University President’s Research Excellence Accelerator Grant.

\bibliographystyle{named}
\bibliography{ijcai24}

\appendix
\newpage
\section{Methodology}
\subsection{Temporal Embedding}
In our proposed TRANS model, we incorporate temporal embeddings to capture the rich temporal relationships in the constructed graph. For irregular visit times, we use Time2Vec to encode time intervals. For temporal edges defined by visit indices, we employ functional time encoding. In \autoref{tvs}, we show the visualizations of these two types of embeddings.
\subsubsection{Time2Vec}
Time2Vec \cite{time2vec} is a model-agnostic representation for time features. It can process irregular time point sampling data. In our model, we use Time2Vec to encode normalized visit times $t$, enabling the model to perceive the intervals between different visits. This helps to better assess the stages of the  patient's illness. Specifically, Time2Vec comprises two parts: periodic and aperiodic components:
\begin{equation}
 \text{T2V}_{2d}(t):= \begin{cases}\omega_i t+\varphi_i, & \text { if } 0\leq i<d. \\ \text{Sin}\left(\omega_i t+\varphi_i\right), & \text { if } d \leq i < 2d.\end{cases}
\end{equation}
where $2d$ represents the dimension of the output vector. The dimensions of the periodic and aperiodic components can be adjusted according to the specific model. $\omega_i$ and $ \varphi_i$ are learnable parameters, and Sin refers to the periodic activation function.
\subsubsection{Functional Time Encoding}
We use temporal edge features to capture the hidden temporal patterns in the topological structures between medical event nodes and visit nodes. For example, it can identify changes in the effects of the same medication at different stages of a patient's treatment, such as the attenuation or amplification of its impact. To achieve a more refined encoding of time indices, we employ functional time encoding proposed by \cite{iclr2020}. This continuous functional mapping $\Phi_d(t)$ from time domain to $\mathbb{R}^{d}$ can be represented as:
\begin{equation}
 \Phi_d(t):=\sqrt{\frac{1}{d}}\left[\cos \left(\omega_1 t\right), \sin \left(\omega_1 t\right), \ldots, \cos \left(\omega_d t\right), \sin \left(\omega_d t\right)\right]
\end{equation}
where the learnable parameters $\omega_i$ correspond to a kernel function that is actually used to quantify the relative timespan.
\subsection{Patient Graph Explainer}
Many attention-based models enhance the interpretability of their predictions by using attention coefficients to assist in analyzing medical events. For models based on GNNs, we adopt a graph explainer framework similar to those proposed for homogeneous graphs \cite{gnnexplainer,pge}. The goal of the explainer is to identify a subgraph of medical events $G_S\subseteq G$ that plays a crucial role in the model's prediction. Specifically, the explainer generates a learnable mask $M_e$ for medical event nodes, which is applied to the original graph $G$. This allows for quantification of the changes in the model's predictions before and after the application of the mask. The greater the change in predictions indicates the higher the importance of the masked nodes. We quantify this change using mutual information:
\begin{equation}
 \text{ MI}\left(\hat{Y}, G\odot M_e\right),
\end{equation}
where \( M_e \) represents the mask, $\odot$ is element-wise multiplication on nodes, and \( G \) is the original graph. In practical implementation, we constrain the sparsity of the graph, meaning \( |M_e| \leq K_m \), \( K_m \) is the maximum node size in the subgraph. Referencing the derivation in \cite{gnnexplainer}, we can obtain the objective function of the explainer as follows:
\begin{equation}
\min _{M_e}-\sum_{i=1}^C \text{I}[\hat{y}_i=y_i] \log P\left(\hat{Y} \mid G\odot \sigma(M_e)\right),
\end{equation}
where $\text{I}[\cdot]$ denotes the indicator function, \( \hat{Y} \) is the output of the original model, and \( y_i \) is the output when the subgraph is used as input. $\sigma$ denotes the activation function.  $C$ is the total number of appeared category labels.
\section{Additional Experimental Setup}
\label{exp_app}
\subsection{EHR Dataset}
\begin{figure}[th]
    \centering
    \includegraphics[width = 0.4\textwidth]{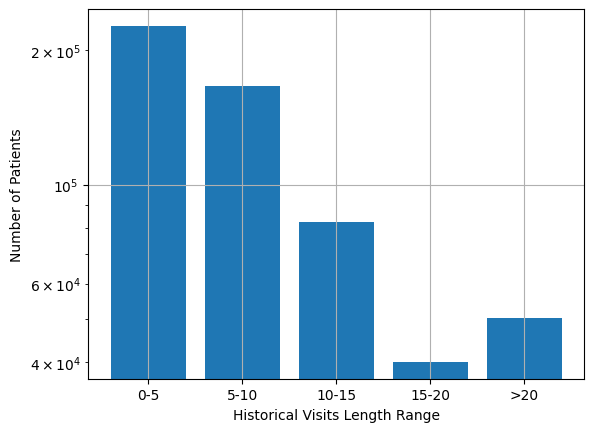}
    \caption{Distribution of visits lengths on the CCAE dataset. `a-b' represents the visit length interval (a, b].}
    \label{fig:stat}
\end{figure}

In this paper, we use MIMIC-III, MIMIC-IV, and CCAE datasets. MIMIC-III and IV are under PhysioNet Credentialed Health Data License 1.5.0\footnote{https://physionet.org/content/mimiciii/view-license/1.4/}. MIMIC-III includes information about patient diagnoses, procedures, medications, and more. MIMIC-IV continues the data structure of MIMIC-III and expands the range and depth of the data.  MarketScan Commercial Claims and Encounters (CCAE) contains the largest number of patients and the most diverse population of the MarketScan data. We identified around 560 thousand distinct patients with Chronic Obstructive Pulmonary Disease (COPD) in the CCAE from 2015 to 2017. We extracted patient data from three source tables: Outpatient Drug (D), Inpatient Admission (I) and Outpatient Services (O). Then we compiled and formulated the raw data into five separate tables that can be easily prepossessed. The details of these tables and demo input data can be found in the Github repository. The statistics on the historical visit lengths of the COPD cohort are shown in \autoref{fig:stat}.

We use the table parsing and data processing pipeline provided in PyHealth \cite{pyhealth2023yang} to construct the three datasets. For the code systems of medical events in the datasets, we use ATC\footnote{https://www.who.int/tools/atc-ddd-toolkit/atc-classification}(level-3) for medication codes. For MIMIC-III, procedures and diagnoses are provided using ICD-9\footnote{https://www.cdc.gov/nchs/icd/icd9cm.htm}, while MIMIC-IV and CCAE use ICD-10\footnote{https://www.cms.gov/medicare/coding-billing/icd-10-codes}. We randomly divide the datasets into the
training, validation, and testing sets in a 0.75:0.10:0.15 ratio.

\subsection{Evaluation Metrics}
We use visit-level precision@k and code-level accuracy@k for the evaluation of the diagnosis prediction task. The term @k denotes the top-k outputs of \( \hat{Y} \) which are ranked by their probability.  We represent the true label and predicted label for patient $t$ as \( Y_t \) and \( \hat{Y}_t \) respectively, where \( Y_t = \{y_1, y_2, ...,y_n\}; y_i\in\{0,1\} \). Visit-level precision@k assesses the predictive performance of a patient's next visit. It can be formalized as:
\begin{equation}
\text{visit-level precision@k}=\frac{\sum_{i=1}^{k}\text{I}\left(\hat{y}_i=y_i\right)}{\min \left(k,\left|Y_t\right|\right)},
\end{equation}
where the numerator represents the number of correct predictions in the top-k prediction, and \( |Y_t| \) denotes the total number of diagnostic codes appearing in the true label of patient $t$. In the paper, we report the average results of visit-level precision@k. Code-level accuracy@k measures the overall accuracy of the model predictions:
\begin{equation}
\text{code-level accuracy@k} = \frac{\sum_{t=1}^P\sum_{i=1}^{k}\text{I}\left(\hat{y}_i=y_i\right)}{\sum_{t=1}^P\left|Y_t\right|},
\end{equation}
where $p$ represents the total number of patients in the dataset.
\subsection{Implementation Details}
For all baselines and our model, we perform ten random runs and report both mean and standard deviation for testing performance. We report the performance on the test set using the best model on the validation set. All experiments are conducted on a machine with a single NVIDIA A100 GPU with 40GB GPU memory. We implement our model with Python (3.10.13), Pytorch (1.12.1 CUDA11.3), torch-geometric (2.3.1), and Pyhealth (1.1.4). We employ AdamW \cite{adamw} as the optimizer for all the models. Next, we introduce the baselines used in our paper.
\noindent \textbf{Baselines:}
\begin{itemize}
    \item Transformer. It is a sequence-based model adapted from NLP. We treat each medical event code as a token and sum the representations of visits, which are then inputted into an MLP-based predictor.
    \item RETAIN. RETAIN \cite{RETAIN} is an RNN-based model that integrates a Reverse Time Attention mechanism. Its implementation is based on PyHealth.
    \item KAME. KAME incorporates medical ontology knowledge of ICD codes into the RNN. For medication codes, we do not introduce any external knowledge.
    \item StageNet includes a stage-aware LSTM module and a stage-adaptive convolutional module. Its implementation is based on PyHealth.
    \item HiTANet. HiTANet \cite{hitanet} is a transformer-based model that treats each visit as a token and incorporates the relative intervals between visits into its attention mechanism.
    \item GCT. It is a classic model for extracting structured information between medical events, but it is not a sequence input model. We apply GCT to extract single visit features, and then input the sequential visit representations into a transformer to obtain the final output.
    \item DDHGNN. \cite{DDHGNN} propose a dynamic heterogeneous graph-based model for Opioid Overdose Prediction. It employs a dynamic heterogeneous graph convolution to learn patient representations and introduces an adversarial disentangler to obtain an enhanced disentangled representation. We adapted this model for diagnostic prediction and omitted the disentangler module in our implementation.
\end{itemize}
We tune the hyper-parameters of baseline models and TRANS by grid search. The search space of hyper-parameters is shown in \autoref{tab:grid}.
\begin{table}[tbp]
    \centering
    \begin{tabular}{c|c}
    \toprule
       Parameters  & Search Space \\
       \midrule
      Learning Rate   & \{1e-2, 5e-3, 1e-3\}\\
      Batch Size & \{64, 128, 256\}\\
      Dropout Rate & \{0-0.6\}\\
    \# of Heads & \{1, 2, 4, 6, 8\} \\
      \# of Layers & \{1, 2, 4, 6\} \\
      Hidden Dimension & \{64, 128, 256\}\\
      Spatial Encoding Dimension$^{*}$ & \{4, 8, 12, 16\}\\
      Temporal Embedding Dimension$^{*}$ & \{8, 16, 32\}\\
         \bottomrule
    \end{tabular}
    \caption{The search space of hyper-parameters. $^*$ TRANS only.}
    \label{tab:grid}
\end{table}

\section{Additional Results}
\subsection{Feature Visualization}
\label{tvs}
In this section, we present the visualization of time and spatial embeddings to provide an intuitive demonstration.
\subsubsection{Temporal Embedding}
\begin{figure}[t]
    \centering
    \begin{subfigure}[b]{0.45\textwidth}
        \centering
        \includegraphics[width=\textwidth]{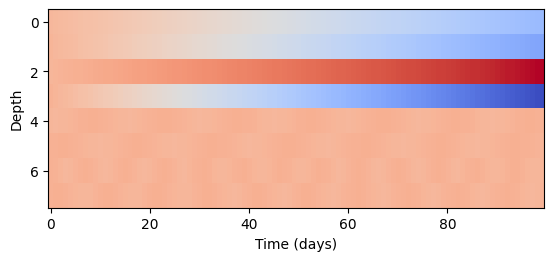}
        \caption{Time2Vec}
        \label{fig:t2v}
    \end{subfigure}
    \begin{subfigure}[b]{0.45\textwidth}
        \centering
        \includegraphics[width=\textwidth]{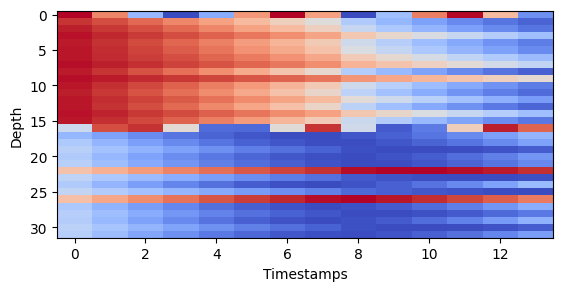}
        \label{fig:fte}
        \caption{Functional Time Encoding}
    \end{subfigure}
    \caption{Visualization of Temporal Embeddings.}
    \label{tsne_app}
\end{figure}
We visualized both Time2Vec and functional time encoding as shown in \autoref{tsne_app}, where different colors represent different values. From the figure, we can observe that both embeddings contain periodic and aperiodic components. On the other hand, we find that compared to Time2Vec, Functional Time Encoding is more sensitive to timespan. Therefore, we use them to encode relatively significant visit time intervals and index-based edge features, respectively.
\subsubsection{Spatial Encoding}
\begin{figure}[th]
    \centering
    \begin{subfigure}[b]{0.45\textwidth}
        \centering
        \includegraphics[width=\textwidth]{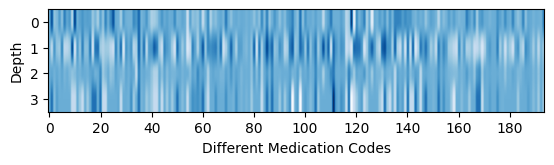}
        \label{fig:ddhgnn_app}
    \end{subfigure}
    \begin{subfigure}[b]{0.45\textwidth}
        \centering
        \includegraphics[width=\textwidth]{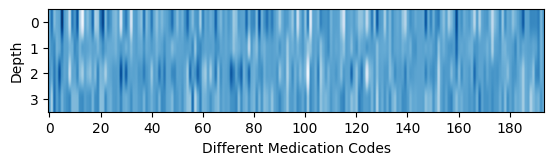}
        \label{fig:ours_app}
    \end{subfigure}
    \caption{Visualization of Positional Encoding.}
    \label{figpe}
\end{figure}
To representatively demonstrate, we show the results of positional encoding for two different patient graphs in \autoref{figpe}. Due to space constraints, we only display the results for medical codes. Positional encoding and structural encoding are capable of representing the structural information of medical event codes within the patient graph. For the same medical event code, global positional encoding can extract its unique positional information across different patients. Meanwhile, local structural encoding can capture certain substructures within a patient's graph, such as ring structures formed by two medication code nodes and three visit nodes. These substructures can be used to deduce high-level patient representations.
\begin{table*}[t]
\resizebox{1.0\textwidth}{!}{
  \begin{tabular}{ll|ccc|ccc}
  \toprule
\multirow{2}{*}{Dataset} & \multirow{2}{*}{Model} & \multicolumn{3}{c}{Visit-Level Precision@$k$} & \multicolumn{3}{c}{Code-Level Accuracy@$k$} \\

& & 10 & 20 & 30 & 10 & 20 & 30\\
\midrule
\multirow{3}{*}{MIMIC-III}&  w/o Seq & $56.85\pm0.33$ & $63.89\pm0.22$ & $73.31\pm0.21$ & $41.13\pm0.35$ & $60.01\pm0.28$ & $70.95\pm0.22$   \\
 & w/o SE &   $57.16 \pm 0.28$ & $64.06\pm0.19$ & $73.50\pm0.20$ & $41.15\pm0.30$ & $60.01\pm0.24$ & $71.00\pm0.19$  \\
  & w/o TE &  $56.73 \pm 0.32$ & $63.70\pm0.20$  & $73.29\pm0.18$ & $41.09\pm0.34$ & $59.91\pm0.23$ & $70.81\pm0.17$ \\

\midrule
 \multirow{3}{*}{MIMIC-IV}&  w/o Seq & $63.85\pm0.28$ & $67.17\pm0.24$ & $75.49\pm0.16$ & $41.61\pm0.26$&   $60.93\pm0.23$  & $72.46\pm0.19$\\
 & w/o SE &   $65.48 \pm 0.24$ & $69.41\pm0.23$ & $77.05\pm0.15$ & $42.70\pm0.21$ & $62.27\pm0.20$ & $73.90\pm0.12$\\
  & w/o TE &    $63.97 \pm 0.27$   &$67.03\pm0.23$ & $75.60\pm0.16$ & $41.75\pm0.25$ & $61.07\pm0.23$& $72.71\pm0.16$\\

\midrule
 \multirow{3}{*}{CCAE}&  w/o Seq &  $76.50\pm0.10$ & $84.80\pm0.12$ & $89.31\pm0.10$ & $71.00\pm0.16$ & $81.94\pm0.11$& $87.18\pm0.12$\\
 & w/o SE & $81.63 \pm 0.08$ & $88.78\pm0.09$ & $92.54\pm0.08$ & $76.15\pm0.09$ & $85.80\pm0.08$ & $90.51\pm0.06$  \\
  & w/o TE &  $76.75 \pm 0.10$ & $85.29\pm0.10$ & $90.35\pm0.08$  & $71.54\pm0.14$ & $82.22\pm0.10$ & $88.32\pm0.09$  \\
\bottomrule
  \end{tabular}}
    \caption{Ablation studies of the TRANS  on three datasets. w/o stands for the
ablated model variant without a specific module.}
    \label{tab:ab_ap}
\end{table*}

\subsection{Length of Historical Visits}
\begin{figure}[th]
    \centering
    \includegraphics[width = 0.4\textwidth]{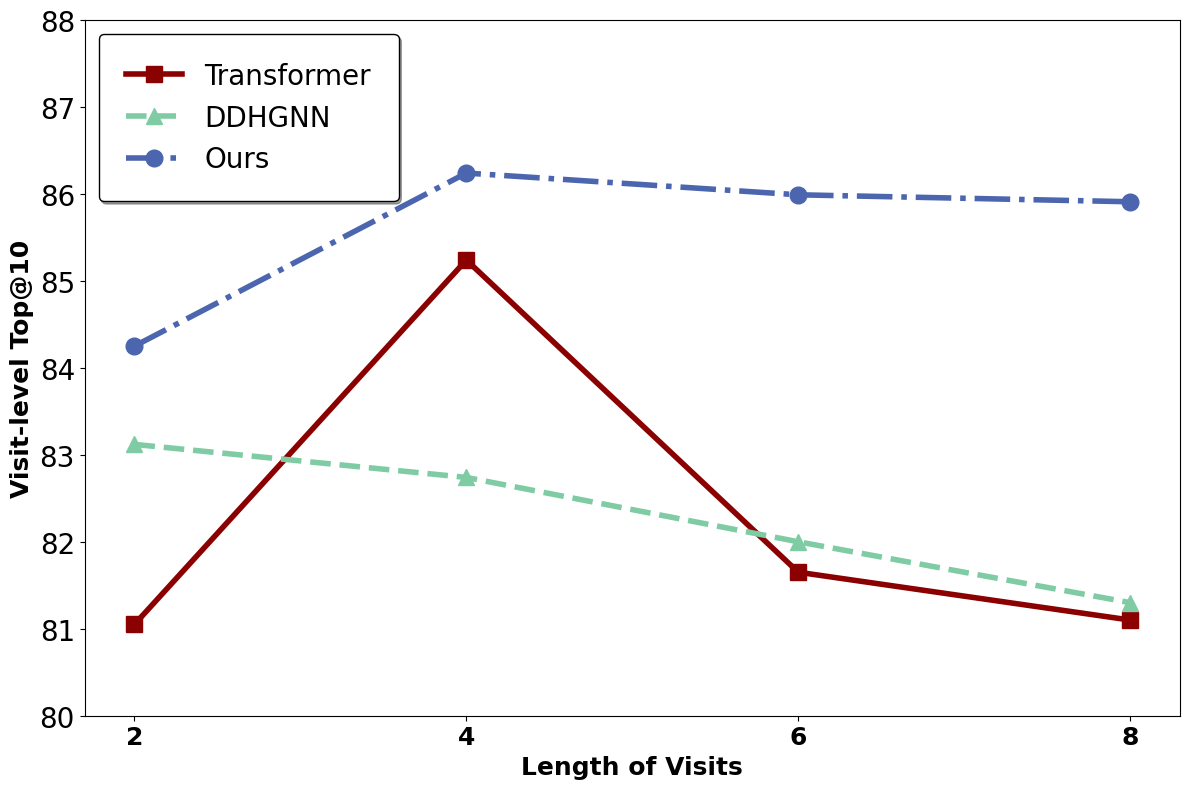}
    \caption{Results for different visit lengths on the CCAE dataset.}
    \label{fig:lohv}
\end{figure}
The length of a patient's historical visits may impact model performance. In this section, we delve into a more detailed examination. As shown in \autoref{fig:lohv}, we test the performance of Transformer, DDHGNN, and our model on patients with visit lengths ranging from 2 to 8 on the CCAE dataset.
All three models perform significantly better on shorter visit sequences compared to the average score across all test samples. This indicates that longer visit sequences are more challenging to predict. Such a result is also consistent with the findings in \cite{anonymous2023graph}. On the other hand, the performance of our model shows a slight decline for longer sequences, but it remains significantly better than both DDHGNN and transformers. We will further explore long-sequence prediction in our future work.

\subsection{Ablation Study}

In \autoref{tab:ab_ap}, we present the performance of various variants of the TRANS model. We observe that compared to structural encoding (SE), temporal information has a greater impact on the model's predictions. Notably, the absence of temporal encoding (TE) has a more significant impact on the model than the absence of sequential information (Seq) from visit nodes. However, in the CCAE dataset, sequential features play a more pronounced role. This is likely because in long visit sequences, relying solely on shallow GNNs means the final visit nodes are unable to capture long-distance dependencies.  Therefore, it is necessary to incorporate sequence features extracted by sequence models into the initial features.
\begin{figure}[th]
    \centering
    \begin{subfigure}[b]{0.22\textwidth}
        \centering
        \includegraphics[width=\textwidth]{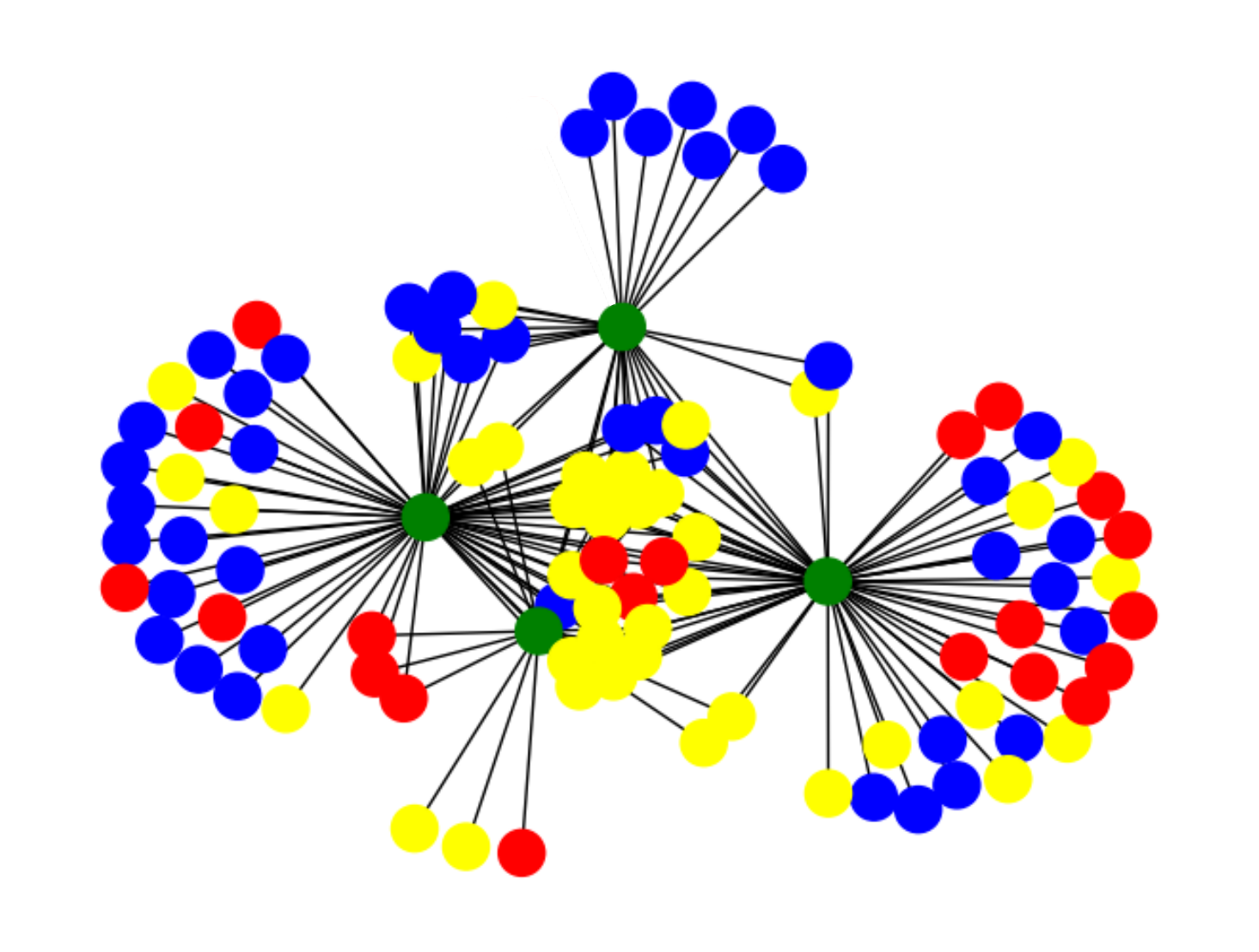}
        \label{fig:image1}
    \end{subfigure}
    \begin{subfigure}[b]{0.22\textwidth}
        \centering
        \includegraphics[width=\textwidth]{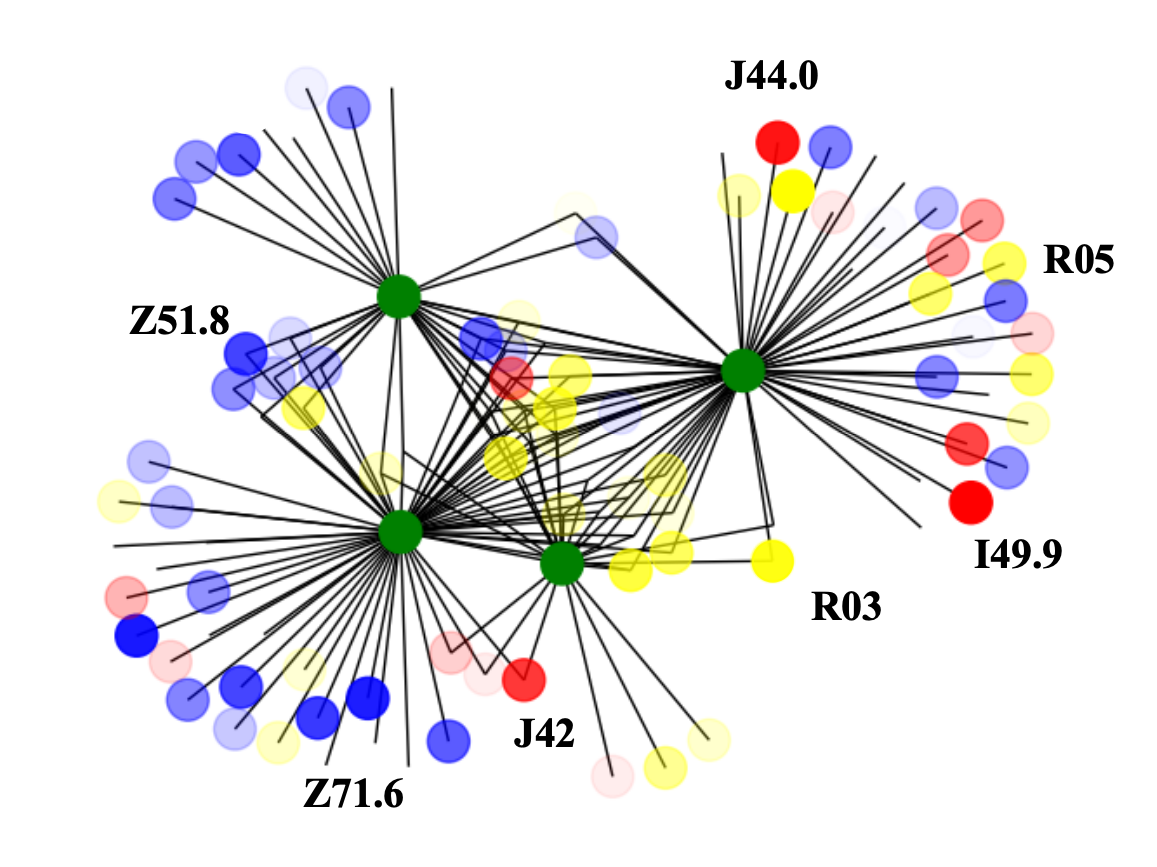}
        \label{fig:ex1}
    \end{subfigure}
    \caption{Initial patient graph v.s. Graph after explainer}
    \label{fig:expl}
\end{figure}
\subsection{Patient Explainer}

\autoref{fig:expl} illustrates the graph derived from a patient's EHR in the CCAE dataset, along with the analysis results using the graph explainer. In this graph, green nodes signify visits, while diagnoses, procedures, and medications are represented by red, blue, and yellow nodes, respectively. The graph structure is visualized using NetworkX, where node transparency in the right-hand figure indicates their importance in the model's prediction. This analysis highlights key diagnostic nodes such as \underline{J41} (chronic bronchitis) and \underline{J44.0} (COPD with acute lower respiratory infection), offering insights into the patient's COPD progression. Additionally, the diagnostic node \underline{I49.9}, indicative of `arrhythmia', suggests a common complication associated with COPD.

\end{document}